\documentclass[submission,copyright,creativecommons]{eptcs}
\providecommand{\event}{Thedu'19} 

\usepackage{underscore}
\usepackage{graphicx}
\usepackage{hyperref}
\usepackage{url}
\usepackage{soul}
\usepackage{color}
\usepackage[dvipsnames]{xcolor}
\usepackage[english]{babel}
\usepackage{algorithmicx}
\usepackage{algpseudocode}
\usepackage{subcaption}

\usepackage[T1]{fontenc}
\usepackage[utf8]{inputenc}

\title{Automating the Generation of High School Geometry Proofs using Prolog in an Educational Context}
\author{Ludovic Font
\institute{\'Ecole Polytechnique de Montr\'eal\\ Montr\'eal, Qc, Canada}
\email{ludovic.font@polymtl.ca}
\and
Sébastien Cyr
\institute{Universit\'e de Montr\'eal\\
Montr\'eal, Qc, Canada}
\email{\quad sebastien.cyr.1@umontreal.ca}
\and
Philippe R. Richard
\institute{Universit\'e de Montr\'eal\\
Montr\'eal, Qc, Canada}
\email{\quad philippe.r.richard@umontreal.ca}
\and
Michel Gagnon
\institute{\'Ecole Polytechnique de Montr\'eal\\ Montr\'eal, Qc, Canada}
\email{michel.gagnon@polymtl.ca}}
\def\titlerunning{Automating the Generation of High School Geometry Proofs using Prolog in an Educational Context}
\def\authorrunning{Ludovic Font, S\'ebastien Cyr, Philippe R. Richard, \& Michel Gagnon}

\begin{document}
\maketitle

\begin{abstract}
When working on intelligent tutor systems designed for mathematics education and its specificities, an interesting objective is to provide relevant help to the students by anticipating their next steps. This can only be done by knowing, beforehand, the possible ways to solve a problem. Hence the need for an automated theorem prover that provide proofs as they would be written by a student. To achieve this objective, logic programming is a natural tool due to the similarity of its reasoning with a mathematical proof by inference. In this paper, we present the core ideas we used to implement such a prover, from its encoding in Prolog to the generation of the complete set of proofs. However, when dealing with educational aspects, there are many challenges to overcome. We also present the main issues we encountered, as well as the chosen solutions.
\end{abstract}

\section{Context}
\label{context}

\subsection{The QED-Tutrix software}
\label{context-qedx}

The QED-Tutrix software~\cite{Leduc2016, Tessier-Baillargeon2015} provides an environment where a high-school student can solve geometry proof problems. One of its key features is that it allows the student to provide proof elements in any order, not limiting them to forward- or backward-chaining. For instance, when solving the simple problem ``prove that a quadrilateral with three right angles is a rectangle'', the student can provide any element of any possible proof, such as a direct consequence of the hypotheses (``if two lines are perpendicular to a third, they are parallel''), a necessary premise for the conclusion (``a rectangle is a quadrilateral that has four right angles''), or anything in between (``the quadrilateral $ABCD$ is a parallelogram''). A second key feature is the tutoring aspect. When the student is stuck is the resolution, the software is able to provide them with relevant messages. In the previous example, if the student entered ``the quadrilateral ABCD is a parallelogram'' and is stuck afterwards, the software identifies that they are working on a proof using parallelogram properties, and will provide them messages such as ``what is the definition of a parallelogram?'' or ``is there a relation between parallelogram and rectangle?''

These features, the flexibility in exploration and the tutoring, are very interesting from a mathematics education perspective, but come with a cost. Indeed, to allow this behavior, the software must know, first, all the various mathematical elements that can be used at any step of any proof for the problem at hand, and second, how these elements are used in the various possible proofs of the problem. This requires, first, a structure that contains all the possible proofs, called the HPDIC graph\footnote{A HPDIC graph contains the Hypotheses, Properties, Definitions, Intermediate results, and Conclusion used in any possible proof for the problem.}~\cite{Leduc2016}, and, second, the generation of such a structure for every problem that is offered in QED-Tutrix~\cite{font2018}. However, even quite simple problems can have a huge amount of possible proofs when considering all the slight variations. This makes the manual generation of HPDIC graphs an extremely tedious and time-consuming task. Furthermore, the software is restricted to proof problems, forbidding the implementation of an interesting aspect of geometry education, the construction problems. In our software, the geometry construction aspect, encapsulated in a dynamic geometry system, is only a tool for the problem resolution.

\subsection{The need for an automated problem solver}
\label{context-need}

\begin{figure}[ht]
  \centering
  \includegraphics[width=1\textwidth]{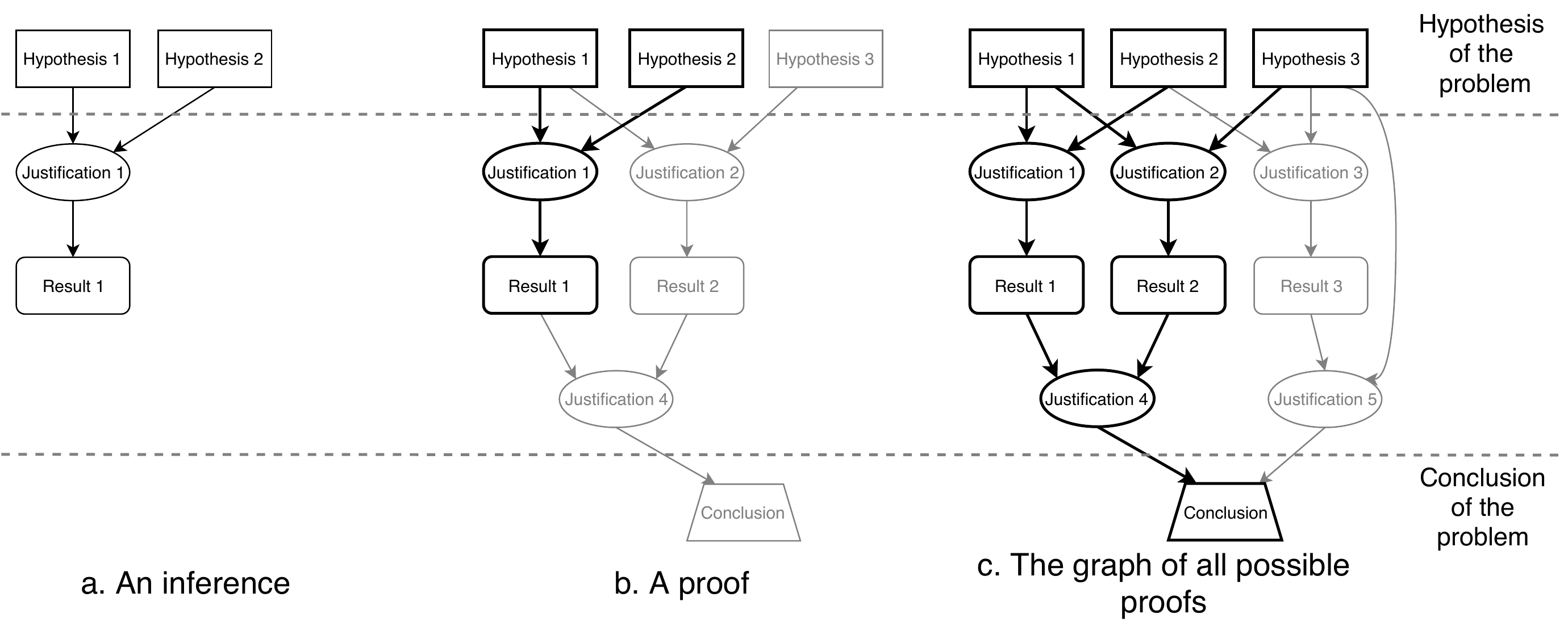}
  \caption{The process of constructing the HPDIC graph}
  \label{HPDIC-constr}
\end{figure}

To avoid having to create the HPDIC graphs by hand, we implemented a tool that, given a problem and a set of mathematical properties, generates the set of all possible proofs. In theory, a mathematical inference is quite easy to model in a computer, as it is essentially the combination of premises (``$ABCD$ is a parallelogram'' and ``$\widehat{ABC}$ is a right angle''), a property (``a parallelogram with a right angle is a rectangle''), and a result (``$ABCD$ is a rectangle), as seen in Figure~\ref{HPDIC-constr}a. This structure is extremely similar to the inference mechanism in logic programming, where a program is a set of facts and rules, and where we infer new facts (the result) based on a rule (the property) and existing facts (the premises). Then, because the result of an inference can be used as the premise of another, a proof is simply a chaining of inferences, starting at the hypotheses of the problem, and reaching the conclusion, as seen in Figure~\ref{HPDIC-constr}b~\cite{duval1995semiosis}. Finally, since the mathematical results can be used in several proofs, we can merge the proofs to create a unique structure containing all the possible proofs for a problem, as in Figure~\ref{HPDIC-constr}c.

Thus, the core of our problem solver seems to be very simple: we create a Prolog fact for each hypothesis and result, and a Prolog rule for each property. However, as we will see in Section \ref{solver}, this simple approach  will not work in practice for most problems, due to technical complications, like combinatorial explosion. Moreover, as shown in Section \ref{challenges}, when confronted with the complexity of human reasoning and behavior in general, and what happens in a classroom in particular, the reality is far more nuanced and difficult to reproduce in a computer.

\subsection{Addition of a problem to QED-Tutrix}
\label{context-addition}
Overall, the addition of a new problem to the QED-Tutrix's database is done in a three-step process.

\paragraph{Creation of the problem:} Quite obviously, the first step is to create the geometry problem, by providing the statement, in the form of a geometrical situation (usually given in a figure, but sometimes in the text as well), the hypotheses, and the desired conclusion. To ensure the functioning of QED-Tutrix, however, one must also ensure that this problem is indeed a \textbf{proof} problem that can be solved by using a chain of inferences. Furthermore, since the automated theorem prover is not able to create new geometrical elements such as lines or circles, it is also necessary to provide the \textbf{super-figure} of the problem. The super-figure is based on the figure that is provided with the problem's statement, to which we add all the geometrical elements that the student could create during the resolution. Therefore, the process of creating the problems must also include predictions on potential construction of new objects. In the current version of QED-Tutrix, the super-figure is provided to the student when he/she begins the resolution process.

\paragraph{Translation in Prolog:} Since we chose Prolog for our automated theorem prover, the representation of the problem must be translated into this language. This is achieved by: 1) identifying the geometrical objects that are present in the super-figure ($AB$ and $BC$ are lines); 2) writing down the hypotheses on these elements ($AB$ is perpendicular to $CD$) and the conclusion of the problem ($ABCD$ is a rectangle); 3) providing some additional information, such as the angles that the student is allowed to use in his/her demonstration, or alternative names for objects (the line $AB$ can also be called $l$). More detail on the encoding is provided in Section \ref{solver-encoding}.

\paragraph{Generation of the HPDIC graph:} Once the Prolog representation of the problem is provided to the software, along with another Prolog file containing all the available mathematical properties, the HPDIC graph, containing all the possible proofs for the problem, can be generated, using the algorithm detailed in Section \ref{solver-proofs}. It is first stored locally in a third Prolog file for testing, analysis and debugging purposes, and, second, if authorized, uploaded directly to QED-Tutrix's problem database. This second part allows for a quick enrichment of the set of problems available in QED-Tutrix.

\section{Related Work}
\label{sec:relatedWork}

This paper presents the advancement of our work on QED-Tutrix. In one of our previous papers~\cite{font2018}, we presented our preliminary results, including a review of research work on tutoring systems and automated theorem proving relevant to this project. Since the basis for our work has not changed, this review is quite similar to the previous one with, however, the addition of some recent work that was not published at the time.

In our work to automatically generate the set of possible proofs to a problem, we have three fundamental constraints that created the need to build our own prover instead of reusing an existing one:
\begin{itemize}
\item the proofs must be \textbf{readable};
\item they must use only properties available at a \textbf{high-school level};
\item there must be a way to handle the \textbf{inferential shortcuts}, i.e. the inference chains that can be deemed too formal by some teachers and are therefore skipped in a demonstration.
\end{itemize}

These constraints direct our search for a way to automatically find proofs. Indeed, there currently exist two general research avenues for geometry automated theorem provers (GATP): algebraic methods and synthetic, or axiomatic, methods. The first one is based on a translation of the problem into some form of algebraic resolution, and the second one uses an approach closer to the natural, human way of solving problems, by chaining inferences.

One of the main goals of the research community in automated theorem proving is the performance. Since synthetic approaches are typically slower, most solvers are based on an algebraic resolution. Algebraic methods include the application of Gröbner bases~\cite{buchberger1988applications,kapur1986using}, Wu's method~\cite{chou1988introduction,wu1979elementary} and the exact check method~\cite{zhang1990parallel}. Practical applications include the recent integration of a deduction engine in GeoGebra~\cite{botana2015automated}, which is based on the internal representation of geometrical elements in complex numbers inside GeoGebra. Other examples include the systems based on the area method~\cite{Chou1996,Janicic2012a}, the full-angle method~\cite{chou1994machine}, and many others. These systems seldom provide readable proofs, and when they do, they are far from what a high-school student would write. Given our readability and accessibility goals, all these systems are not relevant to our interests.

For this reason, we focus on synthetic methods. A popular approach is to use Tarski's axioms, which have interesting computational properties~\cite{braun2017synthetic,narboux2006mechanical}. However, the geometry taught in high-school is based on Euclide's axioms, which are not trivially correlated to Tarski's. Therefore, proofs based on Tarski's axioms are quite inaccessible for high-school students, violating our second constraint. 

A prover that has very similar goals is GRAMY~\cite{matsuda2004gramy}. It is based solely on Euclidean geometry, with an emphasis on the readability of proofs. Besides, it has been developed as a tool for the Advanced Geometry Tutor. It is therefore able to generate all proofs for the given problem. Finally, one of its major strengths is the ability to construct geometrical elements. However, to the best of our knowledge, the source is not accessible, and no work has been done on it since 2004. Furthermore, it does not provide the complete set of proofs.

Another very interesting work is the one of Wang and Su~\cite{wang2015automated}, as it aims at provinding proof for the iGeoTutor, and therefore has the same readability and accessibility at a high school level objectives. In particular, its template-matching algorithm for finding auxilliary constructions is quite promising. However, it requires the use of an external arithmetic engine, and, more importantly, also focuses on finding one proof.

Overall, the very specific needs dictated by the focus on educational interest considerably limit our options. The two only systems with similar goals are GRAMY and iGeoTutor, and they are not suited to our needs. Therefore, we chose to implement our own system.

\section{The problem solver}
\label{solver}

In this section, we present our method to automatically generate the HPDIC graph that represents all possible solutions to a problem.

\subsection{Available data}
\label{solver-data}

\paragraph{Available properties} One of our development goals is to use technology to assist  mathematics education and, more specifically, adapt our software for the different key players within the educational environment (among them, the students, the teachers and the problems' authors). The development of QED-Tutrix is thus an endeavour not only in the field of computer science, but also within the field of mathematics education; therefore, our team is composed of experts in both domains. Although not detailed in this paper, an important part of this work includes collecting, compiling, and organizing the mathematical elements that are used by the software. To date, we have analyzed 18 Quebec high school textbooks ranging from 7th to 11th grade and extracted 1382 mathematical properties and definitions. This large databank of definitions and properties is necessary to properly adapt the software. We call \textit{referential}~\cite{kuzniak2011espace,kuzniak2014espacios} the set of properties and definitions that are allowed or expected in the resolution of a problem. These are known to greatly vary among grades, textbooks, and teachers. Different phrasing might be used from one referential to another even when describing foundational mathematical properties such as \textit{the sum of interior angles in a triangle is 180}\ensuremath{^\circ}. This variability in phrasing must be accounted for to ensure consistency among the properties and definitions used by QED-Tutrix and to allow teachers to choose their innate preferences to be used by their classes. The final cumulative referential extracted from the various school textbooks includes Euclidean geometry, area and volume formulas, metric relations, transformational geometry, analytic geometry and basic vector geometry. We do not attempt to isolate a minimal set of axioms that are used in high school geometry, but instead aim at implementing enough properties and definitions to cover all the material present in high school geometry textbooks. 

\paragraph{Available problems} To learn mathematics, one must solve problems~\cite{brousseau1998theorie}. The tasks of finding, adapting and creating new problems plays an important role in a tutoring system such as QED-Tutrix. Currently, there are approximately fifty problems covering a vast array of mathematical topics from high school courses. These problems have been intentionally selected to allow many possible solutions, each one involving several inferences. In our work, problems are also used as a validation method to ensure a proper encoding and the completeness of the implemented referential. Although those first fifty problems can be used to test the referential, we also need simpler problems to test specific elements of the referential. We thus created smaller problems with one or two inferences per property and definition implemented in our software. Since those simpler problems use a small number of properties or definitions that are related to specific small sub-parts of the referential, we can easily detect errors arising in our proof generator. These errors can be due to some coding mistake or to a missing property or definition within the referential.

\begin{figure}[ht]

    \centering
    \begin{subfigure}[b]{0.45\textwidth}
    	 \textit{``Prove that any quadrilateral with three right angles is a rectangle.''}
    	\includegraphics[width=1\textwidth]{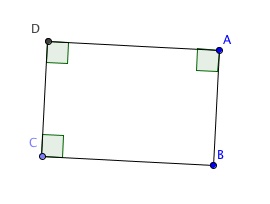}
        
	\caption{Problem statement.}
	\label{subfig:rectangleStatement}
	\end{subfigure}
	\quad
    \begin{subfigure}[b]{0.50\textwidth}
    \centering
    \fbox{\begin{minipage}{20em}
    \small
    hypothese(point(a)).\\
    hypothese(point(b)).\\
    hypothese(point(c)).\\
    hypothese(point(d)).\\
    \\
    hypothese(line([a,b])).\\
    hypothese(line([b,c])).\\
    hypothese(line([c,d])).\\
    hypothese(line([d,a])).\\
    \\
    \underline{hypothese(isAQuad(quad(a,b,c,d)))}.\\
    \\
    \underline{hypothese(angleValue(angle([d],a,[b]),value(90))).}\\
    \underline{hypothese(angleValue(angle([b],c,[d]),value(90))).}\\
    \underline{hypothese(angleValue(angle([c],d,[a]),value(90))).}\\
    \\
    conclusion(rectangle(quad(a,b,c,d))).
    
    usefulAngle([a],b,[c]).
    \end{minipage}}
    \caption{Problem encoding.}
	\end{subfigure}
	\caption{The translation of the rectangle problem from its statement.}
	\label{fig:rectangleTranslation}

\end{figure}

\subsection{Encoding of a problem}
\label{solver-encoding}

The second step is to translate the mathematical problem into a Prolog file. An example of such a translation is provided in Figure~\ref{fig:rectangleTranslation}. The resulting file has several parts:

\paragraph{Implicit hypotheses} The first eight lines provide names to the geometrical objects present in the problem figure. We refer to these as \textbf{implicit hypotheses}, since they are needed for the prover, but are not specified in the problem statement. Furthermore, even though the names of the points are explicit in this case, the Prolog engine needs names for each point, line, etc., even if they are not at all given, neither in the statement nor in the figure. In that case, arbitrary names must be provided (such as P1, P2, etc.). Lines 1 to 4 provide the set of points present in the figure. Then, lines 5 to 8 indicate which point are linked by a line. Here, we have the four lines $(AB)$, $(BC)$, $(CD)$, and $(AD)$ (we do not differentiate segments and lines). If we wanted to add the line through A and C, we would add $hypothese(line([a,c])).$ to the file. The example here is simple enough, but in more complex problems, the encoding of lines can become a delicate issue. Indeed, we must provide, \textbf{as soon as the problem is encoded}, all the points that are on the line. This set of points is, as far as Prolog is concerned, the unique identifier for the line. A direct consequence of this implementation is that it becomes impossible to add, during a proof, new points to a line.

\paragraph{Explicit hypotheses and conclusion} The next four lines provide the hypotheses in a more general sense, meaning the ones that are usually given explicitly in the problem statement, hence the name \textbf{explicit hypotheses}. Here, the statement of the problem, ``Prove that any quadrilateral with three right angles is a rectangle'', with the addition of the figure, provides four hypotheses: there is a quadrilateral named ABCD, and three of its angles, $\widehat{DAB}$, $\widehat{BCD}$ and $\widehat{CDA}$, are right angles. The problem statement also provides the expected conclusion: $ABCD$ is a rectangle, encoded in the second-to-last line.

\paragraph{Auxiliary hypotheses} In many problem resolutions, there is, at some step, the need for the construction of additional elements, such as a new line or point. However, the creation of such elements is a difficult issue. Given that our aim in this project is not actually to solve problems in and of itself, but to ease the process of adding new problems to QED-Tutrix, we chose to require the addition of \textbf{auxiliary hypotheses}, i.e. elements that are not present on the problem statement directly, but are useful in one or several of its resolutions. For instance, in the rectangle problem, one could write a proof using the diagonals of the rectangle. The fact that $(AC)$ is a line would therefore enter in this category. Structurally, auxiliary hypotheses are identical to implicit and explicit hypotheses, it is their origin that differs. When adding these auxilliary elements to the figure of the problem, we create the \textbf{super-figure}.

\begin{figure}[ht]
    \centering
    \includegraphics[width=0.4\textwidth]{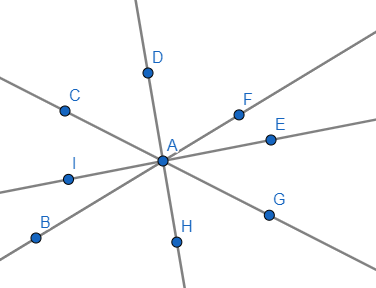}
        
	\caption{A geometrical situation with many angles.}
	\label{fig:multipleAngles}
\end{figure}

\paragraph{Additional elements} The last line in our example is neither a hypothesis nor a conclusion, but an additional information provided to the prover. It can be of two types: useful angles and dictionary items. The useful angles are required because in some geometrical situations, such as the one in Figure~\ref{fig:multipleAngles}, the number of possible angles can become quite huge. Here, there are 8 points around the center A, meaning that there are $8\times 7$ angles. When considering inference such as ``two angles that share a side are adjacent'' and ``two angles whose sum measures 180 degrees are supplementary angles'', the number of possible inferences becomes very impractical. For this reason, to limit the combinatorial explosion that plagues many synthetic automated theorem provers, we chose to impose the following requirement: the problem file must contain the list of all angles on which Prolog is allowed to infer results. In other words, if an angle is not explicitly written in the problem file, no result that depends on this angle will be present in the HPDIC graph. Finally, the dictionary allows the user to provide alternative names for geometrical objects. This is useful in problems where lines, angles, circles, or, more rarely, triangles and quadrilaterals, have alternative names, such as ``line $l$'' or ``rectangle $R$''. This is not needed for the solver, but is important for the tutor software, since the students must be able to enter their sentences with any valid name. 

One should note that, as required by the Prolog language, each constant must have its first letter in lowercase, otherwise Prolog would consider it as a variable. This could lead to some collisions if there is a point named $A$ and a point named $a$, but, in our experience, the conventions used in the statement of high school geometry problems prevent that situation. It is however possible, and usual, that two different mathematical objects are named with the same letter, one uppercase and one lowercase, for instance a circle $c$ and a point $C$, but we avoided collisions by giving an explicit type to each object except points. This means that the point $C$ is named ``c'', and the circle $c$ is named ``circle(c)'' in the Prolog code. 

\begin{figure}
	\begin{algorithmic}[1]
    	\Require $problemFile.pl$
		\State $Hypotheses, ~Conclusion \leftarrow problemFile.pl$
        \State $ToExplore \leftarrow Hypotheses$
        \State $KnowledgeBase.statements \leftarrow Hypotheses$
        \State $KnowledgeBase.inferences \leftarrow \emptyset$

        \Repeat 
        \State $NewPremise \leftarrow ToExplore.pop()$
        \State $NewInferences \leftarrow \Call{InferUsing}{NewPremise, KnowledgeBase}$
        \State $NewResults = \{ Inference.result ~|~ Inference \in       NewInferences \}$
        \State $ToExplore \leftarrow ToExplore \cup NewResults$
        \State $KnowledgeBase.inferences.\Call{add}{NewInferences}$
        \State $KnowledgeBase.statements.\Call{add}{NewResults}$
        \Until{$IsEmpty(ToExplore)$}
        \If{$Conclusion \in KnowledgeBase.statements$}
        \State $Graph \leftarrow \Call{ConstructGraph}{Conclusion, ~KnowledgeBase.inferences$}
        \State \Return $Graph$
        \Else
        \State \Return Error
        \EndIf
        
	\end{algorithmic}
	        ~\\

		\begin{algorithmic}[1]
    	\Function{inferUsing}{Statement, KB} 
    	    \State $Inferences \leftarrow \emptyset$
            \State $CandidateInferences \leftarrow \{ I  ~|~  I \in KB.inferences ~\&~ Statement \in I.premises   \}$
            
            \ForAll {$Inf \in CandidateInferences$}
            \If{$KB.\Call{Prove}{Inf}$}
            \State $Inferences$ = $Inferences \cup \{ Inf \}$ 
            \EndIf 
            \EndFor
            \State \Return $Inferences$
        \EndFunction
	\end{algorithmic}
    \caption{The graph construction algorithm.}
    \label{fig:algorithm-prolog}
\end{figure}

\subsection{Generation of the complete set of proofs}
\label{solver-proofs}

Once the problem is encoded in Prolog, we proceed to the proof generation process. This process follows the algorithm displayed in Figure~\ref{fig:algorithm-prolog}. To summarize, we proceed to a construction of the graph by forward chaining. We maintain a set $ToExplore$, that contains the mathematical results  already known, but from which new results have not been inferred. At the beginning, this set contains only the hypotheses of the problem. We also maintain a knowledge base, that contains statements (hypotheses and inferred results) and the complete inferences. Iteratively, we take one item of the set $ToExplore$ and we ask Prolog to infer everything that can be inferred by using only inferences that have this item in their premises. Each inference found this way is added to the knowledge base (remember that an inference is a property with its premises and its result), and its result added to the set $ToExplore$ and to the knowledge base. We loop until there is nothing new to explore, i.e. Prolog inferred everything using the given hypotheses and properties. Since the set of hypothesis, the set of geometrical elements, and the set of available properties are all finite, the set of possible inferences is also finite. Since the algorithm constructs new inferences without any fallback, we can guarantee that this algorithm terminates. 

Then, we verify if we were able to infer the conclusion. If not, there is an error. Otherwise, we explore the generated graph in backward chaining, starting from the conclusion, and marking an element on the graph only if it can be used to infer the conclusion. This last step is necessary, because the inference engine can infer results that are valid, but useless for the resolution of the problem, i.e. not used in any proof that reaches the conclusion. These results are marked, but not removed from the graph. This will be useful when the student works on the problem and enters such a result. In this situation, the software should not say ``This result is false'', but instead ``This result is valid, but not useful for the problem. Try something else.'' For this behavior to be possible, the ``valid but useless'' elements must remain in the graph and be identified as such.

Since the generated graph is based on mathematical properties, which typically have a reciprocal, there are usually many cycles in the resulting graph. For example, if $ABC$ is a right triangle in $A$, then we can infer that $\widehat{BAC}$ is a right angle. And since $\widehat{ABC}$ is a right angle, we can infer that $ABC$ is a right triangle in $A$. This may be a problem for the calculation of the students' progress in the proof resolution, but it is solved by a clever exploration of the graph. This algorithm, however, is not part of the proof generation process, and is not detailled in this paper. 

In this algorithm, only the \textit{inferUsing} function uses Prolog. The rest of the algorithm is handled by a Python program combined with the QED-Tutrix central database. This database is the core of all our operations, since it contains all the data provided by the input Prolog files (the one that provides the representattion of the problem and the one that contains all the mathematical properties), and all the inferences that have been added and which constitute the generated HPDIC graph.

\subsection{Generation of mathematical results using logic programming}
\label{solver-result}

In the previous algorithm, we abstracted the use of Prolog by a inferUsing fuction, that takes as arguments the knowledge base, representing the set of all results and inferences known by the Prolog engine, and a premise, i.e. a mathematical result that must be used as the premise of the returned set of new inferences. In other words, if the given premise parameter is ``$ABC$ is a right triangle in $A$'', then all the inferences returned will have this statement as a premise. For instance, the inference ``$ABC$ is a right triangle in A, therefore, since a right triangle has a right angle, the angle $\widehat{BAC}$ is right'' is a valid candidate, but ``The angle $\widehat{BAC}$ is right and $ABC$ is a triangle, therefore, since a triangle that has a right angle is right, $ABC$ is a right triangle'' is not, since the given statement is not a premise of the inference, but its result. This requirement is translated into Prolog in three parts. 

\begin{figure}
    \centering
    \fbox{\includegraphics[width=0.7\textwidth]{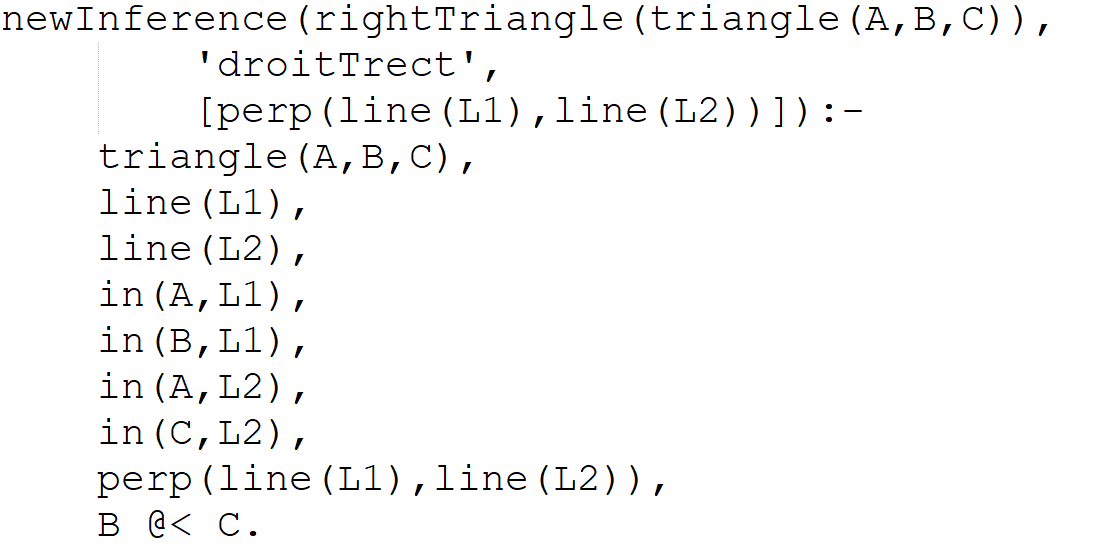}}
    \caption{Translation in Prolog of a property on right triangles}
    \label{fig:property-translation}
\end{figure}

\paragraph{Translation of mathematical properties in Prolog rules} Each property identified during the work presented in Section \ref{solver-data} is encoded into Prolog rules. In Figure~\ref{fig:property-translation}, we translated the property ``a triangle that has a right angle is right''. Let's suppose, for example, that at some point in the solving process, the Prolog knowledge base contains a relation of perpendicularity between two lines $(AB)$ and $(AC)$, and that there exists a triangle $ABC$ with two sides on these lines, encoded by: 
\begin{verbatim}
    perp(line([a,b]),line([a,c])).
    triangle(a,b,c).
\end{verbatim}
In this situation, this rule can be invoked to create the new fact:

\begin{verbatim}
    newInference(rightTriangle(triangle(a,b,c),'rightTrPerp',
        [perp(line([a,b]),line([a,c])]).
\end{verbatim}

\begin{figure}
    \centering
    \includegraphics[width=0.5\textwidth]{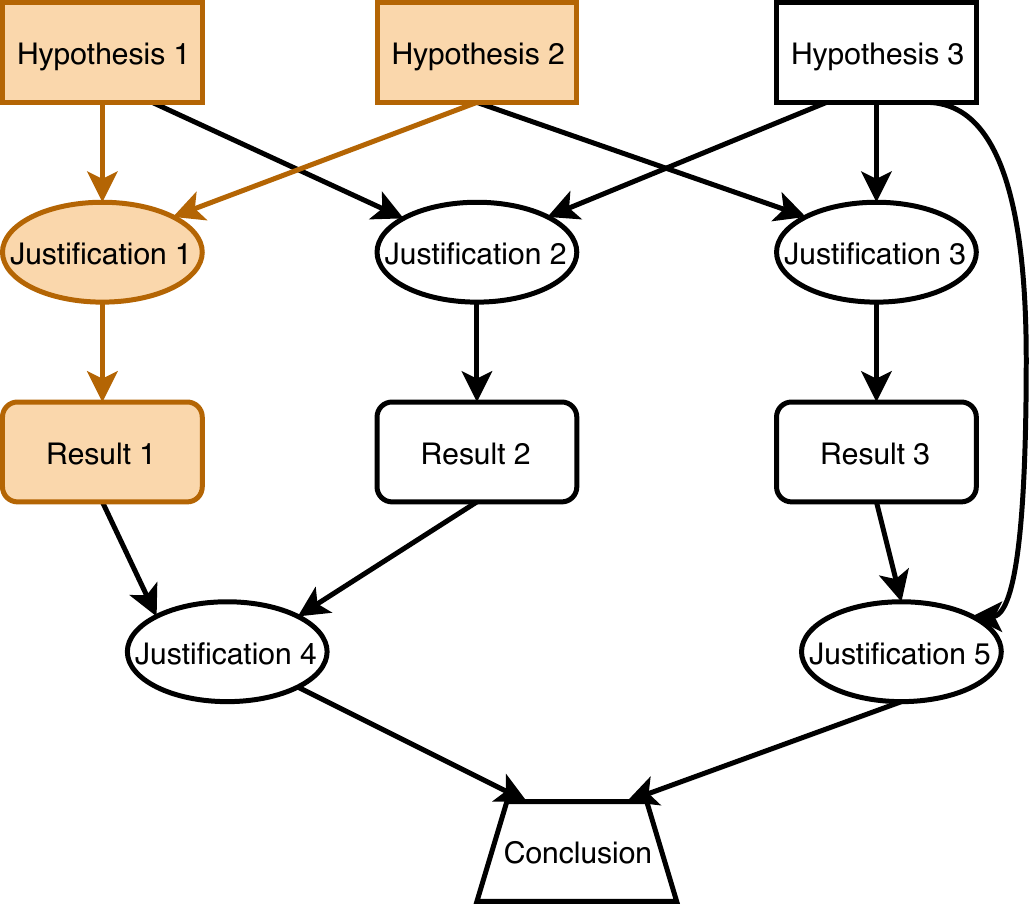}
    \caption{The part of a HPDIC graph encoded by one newInference fact}
    \label{fig:hpdic-prolog}
\end{figure}

This contains the whole inference: the premise ``$(AB)$ perpendicular to $(AC)$''; the property ``rightTrPerp'', which is an encoding for the sentence ``A triangle that has a right angle is right''; and the result ``the triangle $ABC$ is right''. In other words, the knowledge base of Prolog is actually not constituted of simple statements such as ``perp(line($[$a,b$]$),line($[$a,c$]$))'', but of inferences that also contain the \textbf{origin} of the statement, i.e. the property and premises that were used to infer it. In Figure~\ref{fig:hpdic-prolog}, we highlighted the part of an HPDIC graph that is encoded by one of these Prolog facts. 

\begin{figure}
    \centering
    \fbox{\includegraphics[width=0.6\textwidth]{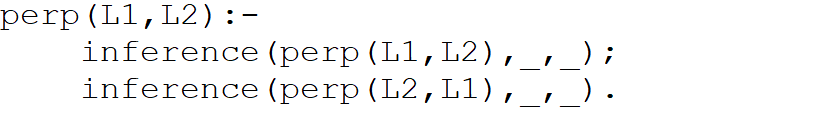}}
    \caption{Link between a statement and an inference}
    \label{fig:property-linking}
\end{figure}

\paragraph{Relation between inference and statement} Since statements such as ``$(AB)$ perpendicular to $(AC)$'' are not encoded as such in the knowledge base, Prolog must have a way to list the available results. Indeed, in Figure~\ref{fig:property-translation}, the second-to-last line checks if the fact perp(line($[$a,b$]$),line($[$a,c$]$)) is true, but this fact is not known directly, only that there exists a newInference(perp(line($[$a,b$]$),line($[$a,c$]$)),_,_). This linking is done by another rule, displayed in Figure~\ref{fig:property-linking}. The translation of a newInference fact into an inference fact is done by the Python wrapper (line 10 in the algorithm). The distinction between the two is crucial to avoid infinite loops in the Prolog inference engine. Furthermore, as seen in the figure, the usage of another rule to obtain the result from the inference allows more flexibility in the order of the elements. Basically, the fact ``$(AB)$ and $(AC)$ are perpendicular'' can stem from an inference whose result is ``$(AC)$ and $(AB)$ are perpendicular''. Both results are obviously equivalent for a human, but not for Prolog. 

\paragraph{Finding an inference using a given premise} Lastly, in the algorithm displayed in Section \ref{solver-proofs}, we restrict the inferences to those that use a given premise. This is done for performance reasons. In some moderately difficult problems, the HPDIC graph can become quite big, making its exploration time consuming. In a previous version of the prover~\cite{font2018}, we used all the available inferences, stored their results, then started again, until no new result is found. This algorithm had the major drawback of duplicating inferences: at step N, the inferences found at step N-1 are still valid, and are therefore found again, increasing the complexity of the algorithm to quadratic on the number of nodes in the final graph. Using the new algorithm makes the complexity linear. 

To achieve this result, we create a Prolog fact for each premise of each mathematical property, linking the two. For instance:
\begin{verbatim}
    usedIn(perp(_,_),rightTrPerp).
\end{verbatim}
indicates that the statement ``perp'', regardless of its argument, is a premise of the rule shown in Figure~\ref{fig:property-translation}. Then, the function inferUsing(NewPremise, KnowledgeBase) used in the algorithm of Figure~\ref{fig:algorithm-prolog} consists in calling Prolog with the query:

\begin{verbatim}
    usedIn(Premise,J),
    newInference(R,J,P).
\end{verbatim}

\noindent that returns in R the result of the inference, J the code of the justification, and P the array of premises, which contains the premise given in argument.

\section{Educational challenges}
\label{challenges}

\subsection{Translation of high school properties into a computable format}
\label{challenges-translation}
In mathematics education, we define different geometric mindsets as the different paradigms GI, GII and GIII~\cite{houdement2006paradigmes}. Our work is a constant back and forth between the latter two (GI concerns  tangible problems of geometry, such as measurements, folding or the use of blocks, which are not the focus of our tutoring system). The natural axiomatic geometry (GII) is best described by the iconic example of the Euclidean geometry. It is a geometry where the reasoning depends on a figure. The component of the inferences are, however, part of the linguistic register, such as the example in Section \ref{solver-data}. A referential in the linguistic register means the statement has semantic and syntactic aspects which can easily be understood by a person. The challenge comes with the translation of these statements into the formal axiomatic geometry (GIII)  of Prolog where the semantics disappears. 

In one of our previous papers~\cite{font2018}, we documented several challenges that arose during the development of the solver. One of the most difficult stems from the fact that the computer must have complete and exact information to infer new results. For instance, when dealing with properties on adjacent or complementary angles, it is crucial to identify exactly each angle. Where most teachers would accept both $\widehat{ABC}$ and $\widehat{CBA}$, interchangeably, in a demonstration, this ambiguity is unacceptable in an automated tool and, if not handled properly, leads to an exploding number of nonsensical results. Therefore, a distinction appears between the process of the solver and what happens in the tutor software. In the solver, the angles are all oriented, and the order of the points is crucial. In QED-Tutrix, however, it depends on the teacher's preferences, but the possibility exists for the teacher to allow both notations. If it is allowed and two results exist on the two opposite angles, for instance $\widehat{ABC} = 120$ and $\widehat{CBA} = 240$, then all four propositions, including $\widehat{ABC} = 240$ and $\widehat{CBA} = 120$, would be accepted. In this kind of problem, though, it is advisable to disallow this behavior, since it uses angles larger than 180 degrees and can lead to some confusion.

This example also touches on another delicate issue, the naming of objects. Here, we mentioned that a teacher might want to allow the students to refer to the angle as $\widehat{ABC}$ or $\widehat{CBA}$. However, if there is a point $D$ on the line $(AB)$, then the angle can also be named $\widehat{DBC}$ (and $\widehat{CBD}$ if allowed). If there is another point $E$ on the line $(BC)$, the number of possible names doubles, with the addition of $\widehat{DBE}$ and $\widehat{ABE}$, and their opposites. All these must be accepted in the tutor software when the student enters a result on the angle. The same is true for lines, since a line containing the three points $A$, $B$ and $D$ can be referred to as $(AB)$, $(BD)$, $(AD)$, etc.

This creates two issues in the implementation. First, Prolog facts representing geometrical objects, such as lines and angles, must contain all their constituting points. For instance, the angle of the previous example is represented in Prolog by:
\begin{verbatim}
    angle([a,d],b,[c,e]).
\end{verbatim}
where the two lists between brackets contain the points on the two half-lines constituting the angle, and the second argument, b, is the center point. This way, this simple fact contains all the information to find the valid names for this angle. In the same way, the line containing $A$,$B$ and $D$ is represented by:
\begin{verbatim}
    line([a,b,d]).
\end{verbatim}
listing all the points on the line. 

Second, since objects can have many names, we must ensure that their representation is unique. This is especially apparent when inferring new results. For instance, if we infer that $(AB)$ is perpendicular to $(BC)$, we only want to infer one Prolog fact representing $(AB) \perp (BC)$, and not $(BC) \perp (AB)$. This required the creation of some rules ensuring the uniqueness of the encoding of a result. In this example, we want the two lines to be alphabetically sorted.

In Section \ref{solver-result}, we explained that each mathematical property is encoded into several Prolog rules. One of them, such as the one in Figure~\ref{fig:property-translation}, is the main one, and contains the restrictions on when that property can be used. Ideally, since a mathematical inference is very close to a Prolog inference, there should be only one such rule per property. However, in reality, it can become necessary to implement several rules for one property. For instance, there is only one property for Pythagoras' theorem: ``In a right-angled triangle, the square of the hypotenuse side is equal to the sum of squares of the other two sides''. There are basically two use cases here. Either we know the value of the two sides and want to calculate the value of the hypotenuse, or we know the value of the hypotenuse and one side and want to calculate the value of the other. This distinction requires the implementation of two Prolog rules. Another example is in the property ``if two of the remarkable lines (bisector, perpendicular bisector, median or altitude) of a triangle are identical, then the triangle is isosceles''. Here, any combination of two remarkable lines is valid, for a total of six possible cases, and, therefore, six Prolog rules.

In these two examples, the need for several implementations is more of a technical need, but in other cases, it happens because the mathematical property is complex and provides many ways to use it. For instance, let us consider the property ``the perpendicular bisectors of the three sides of a triangle intersect in a point that is the center of the circumscribed circle''. There are many ways to use this property: to infer that the three lines intersect at one point; that this point is equidistant to each vertex of the triangle; that the circle whose center is the intersection of the perpendicular bisector and passes through a vertex also passes through the other two. Even though it can be argued that this difficulty lies in the mathematical statement itself, it is a statement that is used in class and, therefore, must be present in QED-Tutrix. Then, when encoding this property in Prolog, we need to consider all the possible use cases and create (at least) one rule for each. In the final result, each inference that uses this property will have the same justification.

\subsection{Differences between a human's and a computer's reasoning}
\label{challenges-reasoning}

As mentioned in \ref{challenges-translation}, the computer is very good with formal proofs, in contrast to students who are better at making mathematical arguments. When solving a geometric problem, students will typically first explore the figure and then reason on it. For example, they will see a triangle, whether or not it is explicited in the problem statement, and will immediately think about what they know about triangles. With the use of informal properties, that we can somehow formalize by a statement such as \textit{I can see it on the figure}, they might determine that the triangle is isosceles. From there the students may start writing their proof. After writing down the aspect related to the isosceles triangle, they might just come back to observe the figure to keep building their reasoning. At some point, they might also use tools (ruler, compass, dynamic geometry software, etc.) to help with this process~\cite{Richard2019Issues}. Overall, this whole process is extremely different from what a computer would do. The algorithm presented in Section \ref{solver} attempts to generate proofs similar to what a student would write, but the process to obtain them is very different. This distinction creates its share of issues.

The main one, and the only one we present in this paper, is the use of different levels of granularity. Indeed, in the mathematics education process, the properties available in the students' referential changes depending on many factors, such as the year of the class, the chapter currently studied, or even the teachers' personal preferences. For instance, when proving a property or theorem, such as ``the sum of the angles in a triangle equals 180 degrees'', students are obviously not authorized to use that property directly, otherwise the proof is trivial. However, after seeing this proof in the classroom, the property can be used to solve exercises. In other words, when considering the HPDIC graph for this problem (``prove that the sum of the angles in a triangle equals 180 degrees''), the resulting property can be seen as a shortcut from the problem hypotheses to the conclusion. And because both the low-level properties used in the complete proof and the high-level, direct property can be useful in class, QED-Tutrix must be able to handle both. Therefore, we identified three ways to implement this duality. The first one is to implement only a basic set of axioms, forbidding the use of such shortcuts, and to scan the generated graph afterwards to add the shortcuts. The second one is to implement only ``advanced'' properties, and, when they appear in the graph, to add the subgraph of their proof to the final graph. Finally, the third way is to implement both, and to generate both paths in the resolution. Even though the handling of shortcuts is still a work in progress at the moment, we chose, for the time being, the third option, to implement both high-level and low-level properties. Further work is needed on both the mathematics education and computer science aspects to determine which solution is the best for the final version.

\section{Limitations}
\label{sec:limitations}

Despite the encountered educational challenges, logic programming is indeed quite adapted to generate the proofs of high school geometry problems. However, it is a lengthy task, and, in its current state, our solver still has several limitations whose correction would require work of varying magnitudes: possible in the current implementation; possible but requiring fundamental changes in the implementation; intrinsically difficult or impossible.  

\paragraph{Possible in the current implementation} Currently, the solver does not handle algebra. For instance, it is impossible to represent the result ``the length of segment AB is three times the length of the segment CD'' without knowing either values. This would require the implementation in Prolog of basic arithmetic operations, but it would be possible, since the use of such results remain in the domain of inferences. Furthermore, we do not handle proofs by contradiction, but, similarly, these kinds of proofs still follow an inferential format. The difference is that the hypotheses represent an impossible situation, and the conclusion is ``we reach a contradiction''. This would require to implement all the numerous possible inferences that result in a contradiction, such as having a triangle with two parallel sides, but it would fit in our solving process. Those two limitations cause another problem concerning an axiomatic referential. Indeed, Euclid's axioms generate most of high school's properties, but most of his work is done by proofs by contradiction which is not currently feasible by our system. On the other hand, Clairaut's adaptation of Euclid's work also generates this geometry and without using proof by contradiction. However, Clairaut uses an algebra system to bypass this difficulty, something that, again, our system cannot do. For this reason, our system requires a much more extensive referential that includes those properties that we cannot generate to be able to solve the problems. This limitation adds another layer of complexity concerning the challenge of the level of granularity discussed in~\cite{font2018}.

\paragraph{Possible with some fundamental change} One of the biggest limitations of our solver is the necessity to provide, beforehand, the whole geometrical situation (the super-figure), including the set of elements that could be useful in a proof. This is very difficult to solve, as the construction of new elements in automated theorem proving in geometry is a whole research domain in itself. However, we envisioned a possible solution. Since QED-Tutrix is an online tool, and that the generation of the possible proofs can usually be done in a matter of seconds, it would be possible to wait for the student to construct a new element in the interface. Then, when constructed, QED-Tutrix sends it to the prover, that attempts to infer new results from this new hypothesis. If we can reach the conclusion, then the hypothesis is indeed useful, and is added, in real time, to the HPDIC graph of the problem. This solution, however, would require a complete change of the interactions between QED-Tutrix and the solver, and comes with its share of issues.
A direct, less profound consequence of this limitation on the possible proofs is that it is impossible to discover, in the middle of a proof, that a line actually passes through a point. This is because it is necessary to provide the complete set of points the line passes through from the beginning when encoding the problem. This limitation could be solved, but it would require a drastic change to the way we represent elements in Prolog.

\paragraph{Intrinsically difficult} A crucial issue comes from the floating-point precision of calculations in the machine. For example, let us assume that there is a right triangle $ABC$, whose dimensions we know. From these dimensions, we can calculate the cosine, sinus and tangent of the angles. This calculation is done with a certain precision. This creates a first problem: what level of imprecision from the student is acceptable? If the cosine of the angle has a value of 0.7654321, is 0.8 acceptable? Is 0.77? Is 0.76? Still, while delicate, this question concerns the tutor aspect, and not the automated proving, and we will not discuss it further. However, there is another issue concerning approximations. In the previous example, now that we know the cosine of the angle, by using the reciprocal of the property, we can calculate the value of the sides of the triangle. This calculation is also done with a certain precision, and if the resulting value is even imperceptibly different from the already known one, then it is considered a new result, that will lead to the calculation of a slightly different cosine, and so on. Hopefully, we were able to solve this issue by only considering the first result. When we already know the value of something, every inference that results in another measurement of that value does not do any calculation, but uses the already known one. In this example, if we already know that the hypothenuse of the triangle is of length 5, then any inference who results in the measurement of that line segment will not even calculate it, but reuse the 5 that has been calculated previously. Lastly, the precision can also create problems when using an inference such as ``two angles are equal if they have the same measure''. Just checking the equality of their measures is not enough, since they could be very slightly different due to rounding. In that case, we allow for a difference of 1\% between the two values. We chose that precision by trial-and-error, with the argument that, in high school geometry problems, two different values that are supposed to be measured will differ by more than 1\%. This is not a mathematically ideal solution, but for our purposes, it is enough.

\section{Other avenues and conclusion}
\label{sec:conclusion}

The future work will consist of several steps. In the short term, we will be completing the implementation of all the identified high school properties in Prolog, and at the same time, add problems to QED-Tutrix as soon as all the useful properties for its resolution are available. Then, one of our major goals is to implement the teacher interface, where they will be able to create new problems, indicate referentials, both for the problem and in general, generate proofs, and obtain statistics of the use of QED-Tutrix by their students (which properties they use, in what order, where they are stuck, etc.). In parallel, we are compiling inferential shortcuts used in class to enrich the possible referentials. Lastly, we are also working on a tool to automate as much of the problem encoding as possible, with, ideally, a software that translates a problem statement in natural language (with, possibly, a figure) into a Prolog file.

Overall, while much work remains to be done, and some limitations of varying degrees exist to our theorem prover, we demonstrated that logic programming is indeed an adequate tool for automating the generation of proofs at a high school level. Furthermore, logic programming offers a high flexibility that allows us to handle the educational challenges stemming for the fundamental differences between the reasoning of a human and the automated theorem proving process. Although challenges remain, the current version is already sufficient to handle an interesting number of high school problems, allowing us to use that prover to create problems for the current version of QED-Tutrix.

\bibliographystyle{eptcs}

\newcommand{\noopsort}[1]{} \newcommand{\singleletter}[1]{#1}

\end{document}